%% file: main.tex
\documentclass[sigconf]{acmart}
\usepackage{xcolor}

\usepackage{enumitem}
\usepackage{multirow}

\AtBeginDocument{
  }

\copyrightyear{2024}
\acmYear{2024}
\setcopyright{rightsretained}
\acmConference[CODS-COMAD Dec '24]{8th International Confernce on Data Science and Management of Data (12th ACM IKDD CODS and 30th COMAD)}{December 18--21, 2024}{Jodhpur, India}
\acmBooktitle{8th International Confernce on Data Science and Management of Data (12th ACM IKDD CODS and 30th COMAD) (CODS-COMAD Dec '24), December 18--21, 2024, Jodhpur, India}
\acmPrice{}
\acmDOI{10.1145/3703323.3703335}
\acmISBN{979-8-4007-1124-4/24/12}

\usepackage[ruled,vlined]{algorithm2e}
\usepackage{graphicx}
\usepackage{hyperref}
\usepackage{algpseudocode}
\usepackage{float}
\usepackage{dblfloatfix}

\begin{document}

\title{Transformers with Sparse Attention for Granger Causality}

\author{Riya Mahesh}
\email{ee21b112@smail.iitm.ac.in}
\orcid{0009-0009-3826-4793}
\affiliation{
  \institution{Indian Institute of Technology Madras}
  \city{Chennai}
  \state{Tamil Nadu}
  \country{India}
}

\author{Rahul Vashisht}
\email{cs18d006@cse.iitm.ac.in}
\orcid{0000-0001-8414-2097}
\affiliation{
  \institution{Indian Institute of Technology Madras}
  \city{Chennai}
  \state{Tamil Nadu}
  \country{India}
}

\author{Chandrashekar Lakshminarayanan}
\email{chandrashekar@cse.iitm.ac.in}
\orcid{0000-0002-3570-7175}
\affiliation{
  \institution{Indian Institute of Technology Madras}
  \city{Chennai}
  \state{Tamil Nadu}
  \country{India}
}

\begin{abstract}
Temporal causal analysis means understanding the underlying causes behind observed variables over time. Deep learning based methods such as transformers are increasingly used to capture temporal dynamics and causal relationships beyond mere correlations. Recent works suggest self-attention weights of transformers as a useful indicator of causal links. We leverage this to propose a novel modification to the self-attention module to establish causal links between the variables of multivariate time-series data with varying lag dependencies. Our Sparse Attention Transformer captures causal relationships using a two-fold approach - performing temporal attention first followed by attention between the variables across the time steps masking them individually to compute Granger Causality indices. The key novelty in our approach is the ability of the model to assert importance and pick the most significant past time instances for its prediction task against manually feeding a fixed time lag value. We demonstrate the effectiveness of our approach via extensive experimentation on several synthetic benchmark datasets. Furthermore, we compare the performance of our model with the traditional Vector Autoregression based Granger Causality method that assumes fixed lag length.
\end{abstract}



\maketitle
\section{Introduction}
Establishing causal links is vital to analyze cause-effect relationships between different natural processes. Causality
analysis finds its application in various domains, spanning from climate discovery \cite{smirnov2009granger},\cite{runge2019inferring} to economics\cite{geweke1984inference}.
The underlying motivation is to understand the patterns and analyze how a change in a particular process can influence another. By identifying causal links, we can potentially mitigate the risks (if any) associated with the ill effects of such changes.
\\
\noindent Identifying potential causal relations from observational time-series data is a challenging task due to the lack of sufficient information on the underlying distribution of the causal model and highly interdependent nature of the that exhibit time lags, auto-correlation, non-linearities and non-stationary characteristics\cite{hasan2023survey}. Several methods using classical machine learning (ML) techniques based on regression \cite{weichwald2020causal},\cite{zhang2022fusing} have been employed to explain causality. These, however, do not disambiguate clearly between correlation and causality. Granger Causality\cite{granger1969investigating} has been instrumental as a pioneering method for identifying causal dependencies in multivariate time series data. The popular framework uses vector autoregressive models to compare the prediction accuracy of the effect variable in the presence and absence of the cause variable\cite{siggiridou2015granger}. 

\noindent Most traditional approaches consider fixed time lags. Our method, however, explicitly tackles the cases of stationary data with long-range dependencies that show random delayed effects due to small, varying lags. We investigate a more realistic setting where the knowledge about the exact time lag with which the past events affect the present is unknown due to such delays. In a real-world scenario, consider a dataset that records crop yield annually over several years along with factors like temperature and precipitation during the monsoon. However, it is very typical that the onset of the monsoon has random delays every year. Traditional models which need a fixed a lag to be specified will fail in this scenario because the lag differs from year to year. On the other hand, using the entire year as the time window has the danger of overfitting, in that, models could use all the time points thereby making it hard to determine causal links. Motivated by this we propose a novel Sparse Attention Transformer (SAT). Our specific contributions are listed below.
\begin{enumerate}[label=\roman*.]
    \item Our sparse attention transformer (SAT) uses a novel sparse attention mechanism which at time $t$ selects the top '$k$' important past time instance within a given window.  In SAT, we don't explicitly specify the lag which enables the model to use the power in the hidden features to learn to select the appropriately lagged time points in a contextual manner.
    \item We experimentally verify performance of SAT on datasets with random delays.
\end{enumerate}

\vspace{-0.25cm}
\section{Problem Setting}
Let $X_t$ = $(X^1_t, X^2_t,..., X^D_t)$ be a set of $D$ variables observed at a time instance $t$. Consider a time-series data observed across $T$ time steps, $X \in R^{T \times D}$ such that $X = \{X_1, X_2,..., X_T\}$ is generated using a single causal model with arbitrary small lags.
Such a dataset can be modeled as 
$X_t$ = $f( X_{t-1-l_1}, X_{t-2-l_2},...., X_{t-k-l_k})$ $\forall$ $t$ where $l_1, l_2, ...,l_k$ are small, variable lags associated with the past time steps.
The task is to generate a causation matrix $\hat{A} \in R^{D \times D}$ such that every entry $\hat{A}_{ij} \in [0,1]$; with 0 indicating the absence and 1 indicating the strong presence of a causal link between $X^i_{t'}$ and $X^j_t$ $\forall$ $t'<t$.   

\section{Background}
\subsection{Causality analysis}
\par \textbf{Multi-variate time series} data is characterised by multiple observations or features recorded at equal intervals of time. The method proposed involves time-series dataset that shows a combination of the following properties:
\begin{itemize}
\item \textbf{Autocorrelaion} is the degree of correlation of a variable at the present instance with its own value in the past.
\item \textbf{Time homogenous} system is where the underlying causal graph remains constant with time.
\item \textbf{Delayed} behavior is demonstrated when the variable is observed at a time much later than its scheduled occurrence.
\end{itemize}

\subsection{Self-Attention in Transformers}
The self-attention layer in a transformer captures the importance weights of every token in a sequence while encoding another token. The attention layer can be visualized as a mapping from queries and key-value pairs to an output\cite{vaswani2017attention}.

\noindent Each token in the input sequence is embedded into a vector. Let the number of words in the sequence be $n$, and each word be represented as a $d$-dimensional embedding. Input is a vector $X \in R^{n \times d}$. The queries, keys and values are obtained by linear transformation of the input sequence as below:

\vspace{0.2cm}
\par $Q = X W_Q$ where $W_Q \in R^{d \times d_k}, d_k$ = dimension of queries

\par $K = X W_K$ where $W_K \in R^{d \times d_k}, d_k$ = dimension of keys

\par $V = X W_V$ where $W_V \in R^{d \times d_k}, d_k$ = dimension of values
\\
\vspace{0.3cm}
The \textit{scaled dot product attention} output is given as
\vspace{-0.3cm}
$$ Attention(Q,K,V) = softmax(\frac{QK^T}{\sqrt{d_k}})V $$ 
Here, the term obtained after softmax operation before multiplying with V is an $n \times n$ attention weight matrix. 

\subsection{Attention for Temporal Causality}
The weights of an attention matrix can be viewed as a measure of the influence one variable has on another and, hence, can be interpreted as an indicator of causal links\cite{rohekar2024causal}. Recent discoveries on causal transformers\cite{melnychuk2022causal},\cite{nichani2024transformers},\cite{shou2024pairwise}, Self-Attentive Hawkes Process\cite{wu2024learning} and improved Granger Causality approaches \cite{zhang2020cause}, \cite{amornbunchornvej2019variable} provide impetus to explore methods based on these mechanisms for causality analysis. 

Previous works involving transformer based mechanisms for causality analysis are on fixed lag datasets. The paper on Causal-Transformer\cite{zhu2024causal} performs independent spatial and temporal causal analysis and then concatenates them to output the final causal analysis results. This method however does not use sparse-attention while training. Sparse attention based transformers are less memory intensive as we calculate lesser number of dot products while
computing causal attention between the variables and hence our method is computationally more efficient. Several methods like Informer\cite{zhou2021informer} and Sparse Transformers\cite{zhang2024sparse} have been developed, but they majorly focus on time-series forecasting and have not been employed for causal analysis. CausalFormer\cite{inproceedings}, uses the ProbSparse attention architecture of an Informer\cite{zhou2021informer} in its time encoder. It separately calculates another attention using causal encoder and feeds it to the decoder to predict the time-series. The informer architecture used here calculates KL divergence between the query-key probability distribution against a uniform distribution. Our mechanism, on the other hand, picks the top-K time-instances contextually by using the column sums of the attention scores without assuming any underlying distribution.

\subsection{Granger Causality}
For stationary, linear time-series data Granger Causality test determines whether a variable X causes another variable Y\cite{siggiridou2015granger}.\\
Consider a D-dimensional vector, spread across T time instances: $X_t = (X_{t}^1,X_{t}^2,....,X_{t}^D)$ $\forall$ $t \in [1,T]$ \\
To check if a variable $X_i$ causes a variable $X_j$, we consider two models for a chosen lag $l$: \\
\textbf{Unrestricted model(U)} is regressing present value of $X_j$ with every variable’s past values including itself
\vspace{-0.2cm}
$$X_{t}^{j} = \sum_{k=1}^{D} (a_{jk,1} X_{t-1}^k + \dots + a_{jk,l} X_{t-l}^k) + u_{t}^j$$
\textbf{Restricted model(R)} is regressing the present value of $X_j$ with every variable's past value excluding $X_i$
\vspace{-0.2cm}
$$X_{t}^j = \sum_{k=1, k \neq i}^{D} (b_{jk,1} X_{t-1}^k + \dots + b_{jk,l} X_{t-l}^k) + e_{t}^j$$
\vspace{-0.2cm}
The Conditional Granger Causality Index is given as
$$\textbf{CGCI}_{X_i \rightarrow X_j} = \ln \frac{{\sigma}^2_R}{{\sigma}^2_U}$$
where ${\sigma}^2_R$ and ${\sigma}^2_U$ are variances of restricted and unrestricted models post regression, with a higher index implying stronger link.

\section{Our Method}
\begin{figure}[htb]
    \centering
    \includegraphics[width=0.6\linewidth]{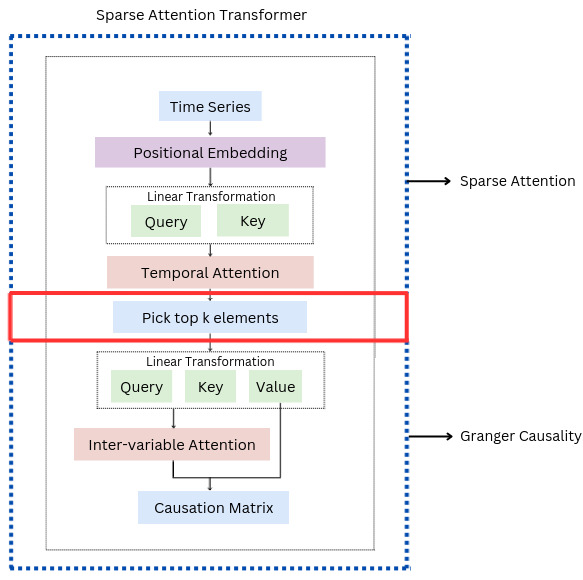}
    \caption{\textmd{Sparse Attention Transformer Model Architecture}}
    \label{fig:enter-label}
\end{figure}

Using the previously established methods as a basis, the concepts of self-attention and Granger Causality are combined to devise a novel algorithm to establish causal relationships. The attention module is modified to select and train on a subset of the original dataset that holds the most important time instances. We introduce Sparse Attention Transformer as shown in [Figure \ref{fig:enter-label}].

\noindent \textbf{Temporal attention}: 
The time-series data $X \in R^{T \times D}$ can be split into samples of input vectors $X'=\{X_{t-1}, X_{t-2},..., X_{t-k}\}$ and the output vector $X_t$ $\forall$ $t>k$ when it demonstrates no delays. The presence of small, random delays can be visualized by modeling $X_t$ as a function of $\{X_{t-1-l_1},....,X_{t-k-l_k}\}$ such that $0 \leq l_i \leq k$ $\forall$ $ i \in [1,k]$. As the data exhibits long-range dependencies in the past, these small delays can be approximated to be upper bounded by the lag value $k$. Hence, the following inequalities can be deduced:

$$t-1-k \leq t-1-l_1 \leq t-1$$
$$\cdots$$
$$t-2k \leq t-k-l_1\leq t-k$$

\noindent Thus, $t-2k \leq t-i-l_i \leq t-1$ $\forall$ $i \in [1,k]$, which indicates the length of the sliding window to be $2k$ to establish the influence of past $2k$ time steps on the present values and pick the $k$ most important time instances. This is analogous to finding the $k$ unknown lags that determine the exact time steps causing the current vector. The above task is achieved by using a temporal attention module.
The time-series data $X \in R^{T \times D}$ can be split into a sequence of $T-2k$ input vectors $\{ \{X_1,X_2,..., X_{2k}\}$, $\{X_2,X_3,..., X_{2k+1}\}$, ..., $\{X_{T-2k},X_{T-2k+1},..., X_{T-1}\} \}$ and their corresponding output vectors $\{ X_{2k+1}, X_{2k+2}, .., X_{T} \}$. Consider an input-output pair at a current time instance $t$. The input vector $X' \in R^{2k \times D}$ is modeled as an input sequence with $2k$ tokens, each having an embedding dimension $D$. The vector $X'$ is then linearly transformed into queries, keys and values as follows:
$$Q_1 = X' W_{Q_1},  W_{Q_1} \in R^{D \times d_{k_1}}$$

\vspace{-0.5cm}
$$K_1 = X' W_{K_1},  W_{K_1} \in R^{D \times d_{k_1}}$$

\vspace{-0.5cm}
$$V_1 = X' W_{V_1},  W_{V_1} \in R^{D \times d_{k_1}}$$
A temporal mask is further applied to prevent the future tokens from being attended. This \textit{mask} is of dimensions $2k \times 2k$ and has its lower triangular and diagonal elements 0 and the remaining as $-\infty$. The temporal attention matrix $A_{t} \in R^{2k \times 2k}$ is obtained as\\
$$ A_t = softmax(\frac{Q_1K_1^T}{\sqrt{d_{k_1}}}  + mask )$$

\noindent Every entry $A_{ij}$ indicates the importance weight $\in$ [0,1] with which token (time instance in this case) $j$ is influencing $i$. The matrix A is a lower triangular matrix with entries in each row summing to 1. This is in accordance with the usual attention weights interpretation, as for a given row, the sum of weights of all the tokens influencing it is 1. 

\noindent We now analyze the physical significance of the column sums of this matrix. The mean of a column $j$ indicates the average influence that time-instance $j$ has on all other time-instances. It is evident that comparing column sums is equivalent to comparing column means as they only differ through a scaling factor of $2k$. Thus, a column with the largest sum corresponds to the time-instance that is most important in predicting the output vector.
\\
Let $S = \{s_1, s_2,..., s_{2k} \}$ be the column sum vector of the temporal attention matrix. We pick $k$ indices with the largest column sums which essentially represent the $k$ most important time instances in predicting $X_t$. Let us denote this index set as $I_k$. We proceed by modifying the input vector $X' = \{X_1, X_2,..., X_{2k}\}$ as $X'' \in R^{k \times D}$ such that $X'' = \{X_i | i \in I_k \}$.\\

\vspace{-0.25cm}
\noindent \textbf{Inter-variable Attention}:
To compute attention between the variables, we start by reformulating the structure of the input. We now transpose $X''$ and model it as an input sequence with $D$ tokens, each having an embedding dimension $k$. This modified input vector

\begin{algorithm}

\SetAlgoLined
\textbf{Input:} Data $X$ with $T$ time samples $X(1), \dots, X(T)$ over $D$ variables\\
\textbf{Output:} Causation matrix $\hat{A}$ of dimension D $\times$ D\\

    $Y \leftarrow (X(2k+1), \dots, X(T))$
    
    $X_{past} \leftarrow X(1) \cdots  X(2k)$ \\
    $\hspace{1.35cm}\vdots  $\\
    $\hspace{1.3cm}X(T-2k)  \cdots  X(T-1)$\\
    $Y' = \phi$ \Comment{List of output vectors of unrestricted model}\\
    $Y'_l =$ $\phi$ $\forall$ l $\in [1,D]$\\ \Comment{List of output vectors on masking $l^{th}$ variable}\\   

    \For{ $i \in [0,T-2k-1]$}
    {   (x,y) = $X_{past}[i]$, $Y[i]$\\
        \textbf{Temporal Attention:}\\
        $Q_1 = x W_{Q_1}$
        \Comment{$x \in R^{2k \times D}, W_{Q_1} \in R^{D \times d_{k_1}}$}\\

        $K_1 = x W_{K_1}$
        \Comment{$x \in R^{2k \times D}, W_{K_1} \in R^{D \times d_{k_1}}$}\\
        $mask$ M $\in R^{2k \times 2k}$\\
        $M_{ij} =$ $ -\infty $ $ \forall$ $i<j$\\
            \hspace{0.9cm} $0$ otherwise\\~\\
        $A_t = softmax(\frac{Q_1K_1^T}{\sqrt{d_{k_1}}}  + M )$
        \Comment{$A_t \in R^{2k \times 2k}$}\\~\\
        
        \For{$l \in [1,2k]$}
        {
            $s_l = ~\sum_{j=1}^{2k} A_{jl}$
        }
        $I_k =$ Set of $k$ indices with largest $s_k$ values\\
        $x' = ~\{ x(p)$ | $p \in I_k \}$ \\
        $x'' = (x')^T$
        \Comment{$x'' \in R^{D \times k}$}\\
        \textbf{Attention between variables:}\\
        \textbf{Unrestricted model} (Training Phase)\\
        $Q_2 = x''  W_{Q_2}$
        \Comment{$ W_{Q_2} \in R^{k \times d_{k_2}}$}\\

        $K_2 = x''  W_{K_2}$
        \Comment{$ W_{K_2} \in R^{k \times d_{k_2}}$}\\

        $V_2 = x''  W_{V_2}$
        \Comment{$ W_{V_2} \in R^{k \times 1}$}\\
        $ A_d = softmax(\frac{Q_2K_2^T}{\sqrt{d_{k_2}}})$ 
        \Comment{$A_d \in R^{D \times D}$}\\~\\
        $y' = A_d V_2 $ 
        \Comment{$y' \in R^{D \times 1}$}\\
        Backpropagate $y'$ w.r.t $y$ using $MSE$ loss\\
        $Y' \gets Y' \cup \{y'\}$\\

        \textbf{Restricted model} (Evaluation phase)\\
        \For{$l \in [1,D]$}
        {   
            $Mask ~ x''_l ~ \forall ~ k$ time steps\\
            Pass through the trained model $\rightarrow y'_l $\\
            $Y'_l \gets Y'_l \cup \{y'_l\}$ 
            
        }
    }
    
\For{$l\in[1,D]$}
{       
Find ${\sigma}^2_{l}$ using $Y'$ and $Y$\\
\Comment{${\sigma}^2_{l}$ is variance in $l^{th}$ variable in unrestricted model}

Find ${\sigma}^2_{l1},\cdots,{\sigma}       ^2_{lD}$ using $Y'_l $ and $Y$\\

\Comment{${\sigma}^2_{lm}$ is variance in $m^{th}$ variable when $l^{th}$ is masked}\\
$\hat{A}_{lm} = \ln\left(\frac{{\sigma}^2_{lm}}{{\sigma}^2_{m}}\right) \quad \forall m \in [1,D]$}

$\hat{A} = \frac{\hat{A}}{\max\limits_{i,j} \hat{A}_{i,j}}
$
\caption{Sparse Attention Transformers for Granger Causality}
\label{alg1}
\end{algorithm}
$X''' = (X'')^T \in R^{D \times k}$ is linearly transformed into queries, keys and values as follows:
$$Q_2 = X''' W_{Q_2},  W_{Q_2} \in R^{k \times d_{k_2}}$$
\vspace{-0.56cm}
$$K_2 = X''' W_{K_2},  W_{K_2} \in R^{k \times d_{k_2}}$$
\vspace{-0.56cm}
$$V_2 = X''' W_{V_2},  W_{V_2} \in R^{k \times 1}$$

On performing self-attention between queries and keys of this input in a similar way, we obtain the attention matrix $A_{d} \in R^{D \times D}$ capturing the dependencies of the D variables over each other as:
$$ A_d = softmax(\frac{Q_2K_2^T}{\sqrt{d_{k_2}}})$$

\vspace{-0.15cm}
The matrix $A_d$ is multiplied with $V_2$ to generate the output vector $Y \in R^{D \times 1}$ that is backpropagated with respect to $X_t$ to minimize the loss function.\\

\vspace{-0.25cm}
\noindent \textbf{Granger Causality using Attention:} The transformer model trained using the above method is now deployed at the inference stage on the given dataset, masking the effect of every variable one at a time. This is achieved by using a mask before applying softmax while computing scaled dot-product attention matrix $A_d$. We apply a $D \times D$ mask with entries of the $i^{th}$ column as $-\infty$ and others as 0 to mask the effect of the $i^{th}$ variable. Hence, new predictions for $X_t$ are obtained with each column being masked individually, keeping the parameters of the model learned during training constant. We now compute the standard deviation of errors for the unrestricted model (predictions obtained during the training phase) and restricted model (predictions obtained during the inference stage by masking variables one by one) and calculate the Granger Causality Index of every variable with past instances of every other variable. We normalize the obtained Granger Causality matrix by dividing each entry with the maximum value in the matrix, producing a final causation matrix as shown in Algorithm \ref{alg1}.

\section{Experimental Setup}
\subsection{Dataset}
We modelled our data generation process based on the framework presented in the Causality 4 Climate (C4C) challenge as part of the NeurIPS 2019 competitions track\cite{runge2020causality}. The data is generated by a function that given the number of variables and maximum lag, creates a graph with probabilistic edges, where weights (linear coefficients) and delays are uniformly sampled, and the data is iteratively accumulated over time steps. The performance of the model is evaluated on such time-homogenous linear datasets with endogenous variables. We  generate 20 different datasets each with 150 time-steps for every group. We also alter the number of variables, to obtain a total of 80 datasets across 4 groups. The maximum lag dependency in the synthetically generated data is 10. Hence, we use a window size of 20 ($2k$) and pick the top 10 ($k$) elements in our transformer model. 
\subsection{Training Details}
Two single-head self-attention layers are used for capturing time instance dependencies and relations between the variables respectively. The input vector is passed through the positional embedding layer to preserve the temporal order of the observed variables. Adam optimizer\cite{kingma2014adam} with Mean Squared Error Loss function is used to train SAT. 

\begin{table}[htb]
\begin{tabular}{c|c|c|c|c}
	\toprule
	\multirow{2}{*}{No. of variables} & \multicolumn{2}{c|}{Our method} & \multicolumn{2}{c}{VAR Granger Causality} \\
	\cmidrule{2-5}
	& AUC-ROC & F1 score & AUC-ROC & F1 score \\
	\midrule
	4 & 0.77 & 0.72 & 0.47 & 0.63 \\
	5 & 0.71 & 0.69 & 0.49  & 0.51 \\
	6 & 0.70 & 0.63 & 0.67 & 0.66 \\
	 10 & 0.78 & 0.65 & 0.63 & 0.62 \\
	\bottomrule
\end{tabular}

  \caption{\textmd{Comparison of mean AUC-ROC and F1 scores across the three groups of data samples with D=4, 5, 6 }}
\label{tab:freq}
\end{table}

\vspace{-1cm}
\begin{figure}[htb]
    \centering
    \includegraphics[width=0.4\linewidth]{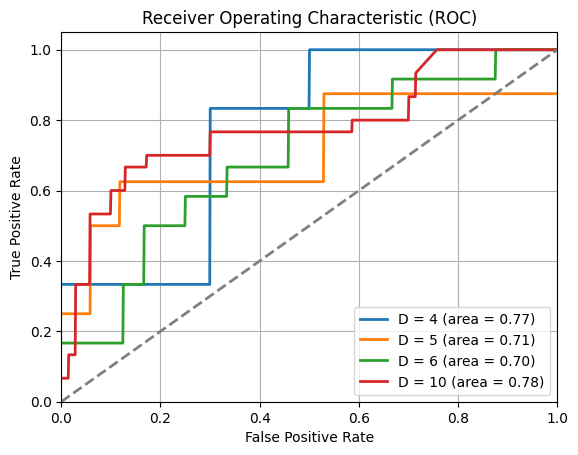}
    \caption{\textmd{ROC plots across four groups of data samples- D=4, 5, 6, 10; D indicating the number of variables in the time-series}}
    \label{fig:roc}
\end{figure}

\vspace{-0.75cm}
\section{Results and Discussion}
We choose the AUC-ROC measure and F1 scores between the causation matrix and ground truth causality matrix as metrics to evaluate performance across various datasets, grouped into sets based on the number of variables in Table \ref{tab:freq}. The ROC plots generated using our method for each of the cases is reported in Figure \ref{fig:roc}.  Vector Autoregression (VAR) based Granger Causality\cite{siggiridou2015granger} with the same window size of 20 is used as a baseline algorithm for our analysis. For the chosen evaluation metrics, mean scores across all the 20 datasets for each group obtained using our method and the baseline are compared. We show that Sparse Attention Transformer performs significantly better than the baseline method.

\section{Conclusion}
In this work, we introduce Sparse Attention Transformer model for Granger Causality. We address the problem of the presence of random delayed effects in time series data and our model successfully identifies the importance weights associated with every variable in predicting the others. Our method outperforms the traditional Granger Causality approach for the synthetic datasets considered. Future work includes extending the method to non-linear as well as non-stationary datasets and explaining the exact mathematical relation of multi-head attention weights with causality coefficients.

\newpage

\bibliographystyle{ACM-Reference-Format}
\input{main.bbl}

\end{document}